\documentclass[conference]{IEEEtran}
\IEEEoverridecommandlockouts
\usepackage{url}
\usepackage{cite}
\usepackage{amsmath,amssymb,amsfonts}
\usepackage{algorithmic}
\usepackage{graphicx}
\usepackage{textcomp}
\usepackage{xcolor}
\def\BibTeX{{\rm B\kern-.05em{\sc i\kern-.025em b}\kern-.08em
    T\kern-.1667em\lower.7ex\hbox{E}\kern-.125emX}}

\usepackage{microtype}
\usepackage{subfigure}
\usepackage{booktabs}
\usepackage{tabularx}
\usepackage{multirow}
\usepackage{array}
\usepackage{makecell}
\usepackage{enumitem}
\usepackage{placeins}
\usepackage{float}
\usepackage{threeparttable}

\newcolumntype{R}{>{\raggedleft\arraybackslash}X}
\newcolumntype{L}{>{\raggedright\arraybackslash}X}
\newcolumntype{C}{>{\centering\arraybackslash}X}

\newcommand{\norm}[1]{\left\lVert#1\right\rVert}

\DeclareUnicodeCharacter{2212}{-}

\begin{document}

\title{Deep Learning for Automatic Quality Grading of Mangoes: Methods and Insights}

\author{\IEEEauthorblockN{Shih-Lun Wu, Hsiao-Yen Tung, and Yu-Lun Hsu}
\IEEEauthorblockA{\textit{Dept. of Computer Science and Information Engineering} \\
\textit{National Taiwan University}\\
Taipei, Taiwan \\
\texttt{b06902080@csie.ntu.edu.tw, \{b06401036, b06401078\}@ntu.edu.tw}}
}

\maketitle

\begin{abstract}
The quality grading of mangoes is a crucial task for mango growers as it vastly affects their profit. However, until today, this process still relies on laborious efforts of humans, who are prone to fatigue and errors. To remedy this, the paper approaches the grading task with various convolutional neural networks (CNN), a tried-and-tested deep learning technology in computer vision. The models involved include Mask R-CNN (for background removal), the numerous past winners of the ImageNet challenge, namely AlexNet, VGGs, and ResNets; and, a family of self-defined convolutional autoencoder-classifiers (ConvAE-Clfs) inspired by the claimed benefit of multi-task learning in classification tasks. Transfer learning is also adopted in this work via utilizing the ImageNet pretrained weights. Besides elaborating on the preprocessing techniques, training details, and the resulting performance, we go one step further to provide explainable insights into the model's working with the help of saliency maps and principal component analysis (PCA). These insights provide a succinct, meaningful glimpse into the intricate deep learning black box, fostering trust, and can also be presented to humans in real-world use cases for reviewing the grading results.
\end{abstract}


\section{Introduction}\label{sec:intro}

Mangoes are a lucrative fruit widely grown in tropical and sub-tropical regions of the world. 
Its enticing aroma, flavorful pulp, and high nutritional value attract numerous mango lovers from worldwide, contributing enormous economic benefits to mango growers and exporting countries. 
It is worth emphasizing that the economic value of a mango fruit depends heavily on the aesthetics of its appearance; the best-looking ones are reserved for export, the lesser ones for domestic consumption, and the worst ones for further processing to make canned fruit or jam.
However, the quality grading of mangoes is a laborious process which, up until now, almost fully relies on human inspection. 
This time-consuming process not only shortens the profitable shelf life of the fresh fruits, but is also prone to human errors that could lead to losses. 

Therefore, this work, coupled with the Taiwan AI CUP 2020 competition, strives to bring the tried-and-tested deep learning technology in computer vision, namely the various convolutional neural networks (CNNs) \cite{lecun1998gradient, krizhevsky2012imagenet, simonyan2014very, he2015deep}, to the rescue of mango growers, helping them finish the grading task accurately and effortlessly.

\begin{figure}[]
 \centerline{
 \includegraphics[trim={0 0 0 0},clip,width=\linewidth]{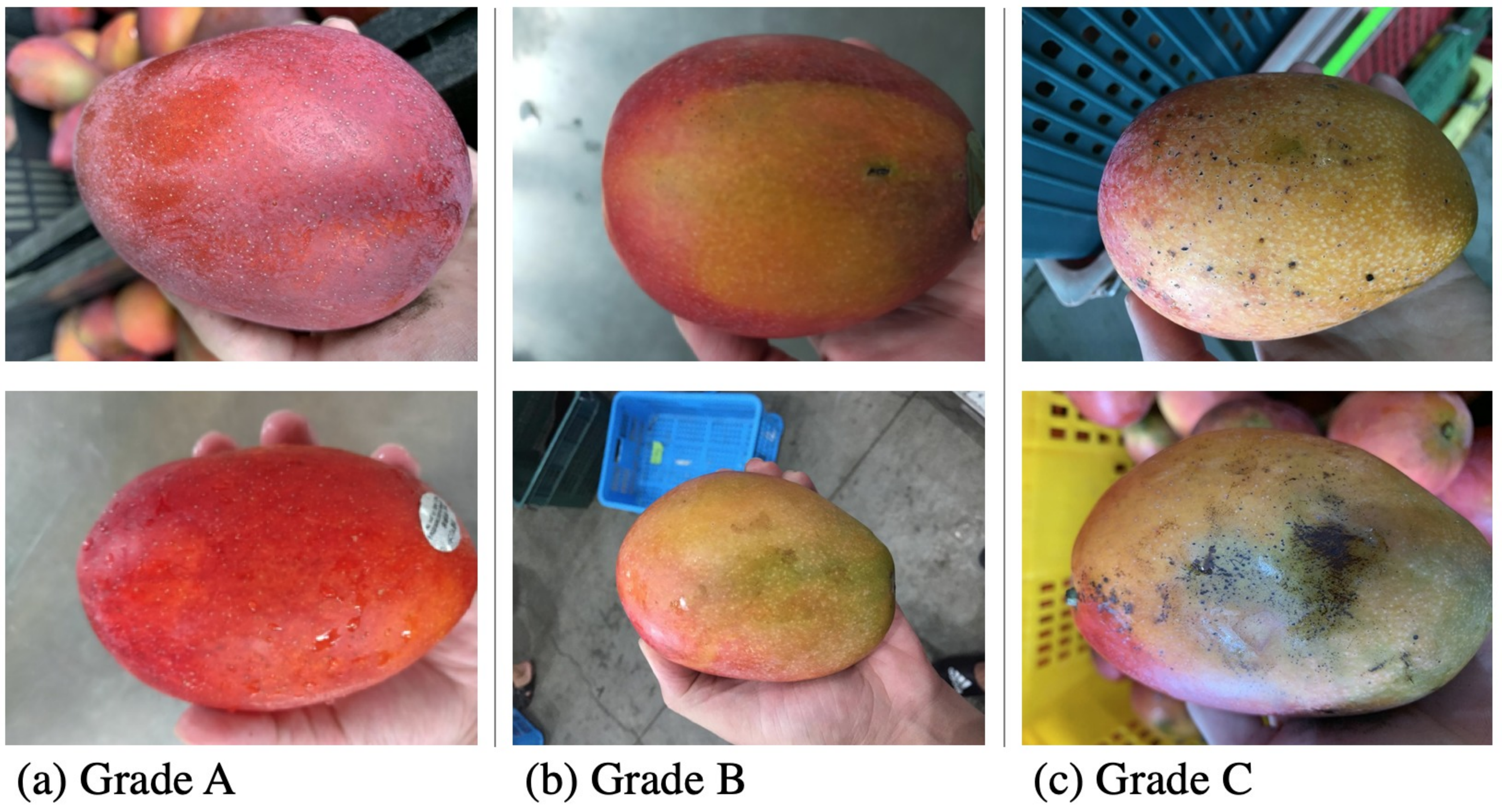}}
 \caption{Some sample images from AICUP2020, the dataset used in our work. Please note that the images possess high variance in background, lighting conditions, and the distance of target mangoes.}
 \label{fig: sample-imgs}
\end{figure}

The challenges of applying machine learning to diversified domains often lie in ensuring the quality of collected data, and opting for the correct suite of existing tools with some task-specific tweaks, rather than innovating brand-new learning algorithms or network architectures; our case is no exception.
The dataset adopted in this work consists of 6,400 images of single mangoes, each labeled with a quality grade of either A, B, or C. However, the photos are taken casually by humans in mango processing plants, leading to issues such as noisy background, varying distance and position of target mangoes, and diverse lighting conditions (see Figure \ref{fig: sample-imgs}). To tackle these, we employ a series of data preprocessing techniques (see Section \ref{sec:data-prep}) to enhance the data quality, one prominent effort being to remove most of the irrelevant background with the help of Mask R-CNN \cite{he2017mask} fine-tuned on our manual annotations of the target mangoes' boundary in the images.

The deep learning models selected for our classification task are all past winners, namely, AlexNet \cite{krizhevsky2012imagenet}, VGGs \cite{simonyan2014very}, and ResNets \cite{he2015deep}, of the ImageNet Large Scale Recognition Challenge (ILSVRC) \cite{russakovsky2015imagenet}, a 1000-class classification task containing more than a million images.
Furthermore, as \textit{transfer learning} \cite{shin2016deep} has been shown effective in carrying knowledge learned from general-domain, large-scale datasets to specific domains, where the amount of data available is limited, we make use of the ImageNet pretrained weights provided by the \texttt{torchvision} package\footnote{\url{pytorch.org/docs/stable/torchvision/models.html}}. 
In addition to the renowned models mentioned above, being motivated by the benefit of \textit{multi-task learning} to classification tasks shown in \cite{zhang2016augmenting}, we also attempt to augment the CNN classifier with a convolutional autoencoder jointly optimized during training.
The rationale of considering such networks is as follows: \textit{1)} the presence of autoencoder forces the network to remember essential information for reconstruction when extracting features for classification, thereby having a regularizing effect; and, \textit{2)} the latent features learned by the network could benefit other downstream tasks as they contain the compressed information for reconstruction; that is to say, we may keep the autoencoder intact and merely place a new classifier part for another related task, e.g., the defect type classification.
The two aforementioned types of networks are dubbed ``single-task CNNs'' and ``convolutional autoencoder-classifiers'' respectively; and, will have their architectural and training details elaborated in Sections \ref{sec:e2e-cnn} and \ref{sec:conv-ae}.

In the case of machine learning applications, oftentimes, being able to provide the reasoning behind the model's decisions is of equal significance as achieving high accuracy. 
Exploiting the model's ``explainability'' not only helps us gain succinct insights into the model's intricate decision process, but also fosters users' trust in the deep learning black box \cite{adadi2018peeking}.
In this light, in our experiments and discussions (Section \ref{sec:exp}), besides presenting the performance of our proposed solutions, we will also take a deeper look into the model's working.
Such measures include examining the focus of the model during prediction with the help of \textit{saliency maps} \cite{Simonyan2014deep}; and, discovering how mangoes of different quality grades are differentiated in the latent feature space via \textit{principal component analysis} (PCA) \cite{hotelling1933analysis}.
By adopting such techniques, the automatic grading system can provide human overseers with more information along with the predictions for reviewing the grading results.

\section{Related Work}\label{sec:related-work}

The recent decade has seen a sizeable body of research on agricultural applications of deep learning techniques \cite{kamilaris2018deep}, more specifically, the various CNN-based architectures. 
However, the majority of works investigated land-use classification \cite{luus2015multiview, lu2017cultivated} or crop type detection \cite{rebetez2016augmenting, kussul2017deep} from aerial/satellite images; and, fruit counting \cite{sa2016deepfruits, bargoti2017deep, chen2017counting, koirala2019deep} or weed detection \cite{potena2016fast, mccool2017mixtures} from images taken from closer distances. 
Another series of efforts paid attention to plant disease detection \cite{ferentinos2018deep, ma2018recognition, toda2019convolutional}, but all of them examined plant leaves rather than fruits.

Notwithstanding, if we narrow down the scope of the study to the quality grading of fruits, there were almost no deep learning-based solutions proposed. 
The two previous works concerning the grading of mangoes \cite{vyas2014quality, nandi2016machine} involved no deep learning, and relied substantially on the size, shape, and color features computed from meticulously-taken images; as well as completely rule-based grading criteria.
A more recent research on other types of fruits \cite{bhargava2020automatic} though experimented with several classical machine learning methods (e.g., KNN, SVM, etc.), still put great emphasis on feature engineering rather than approaching the task in an end-to-end fashion. 

Therefore, our work represents the pioneering effort to bring the tremendous success of deep neural networks to this realm of research; moreover, in addition to employing the existing deep networks, we go one step further to provide some valuable insights into the model's decisions (see Sec. \ref{subsec:insights-model}) to justify that, with deep learning, comprehensive feature engineering and meticulous photographing are no longer necessary.

Regarding multi-task learning, \cite{zhang2016augmenting} and \cite{le2018supervised} have demonstrated its benefit to classification tasks; the former work presented the performance gain on the large-scale ILSVRC dataset, while the latter focused on providing theoretical explanations of an autoencoder's assistance in generalization and regularization. Besides, a work in the medical domain \cite{shvets2018angiodysplasia} also revealed the effectiveness of using convolutional autoencoders, accompanied by ImageNet pretrained network components, for human disease detection and localization. These works serve as our motivation to propose solutions based on convolutional autoencoder-classifiers (see Section \ref{sec:conv-ae}) to see whether they are competent for our task.

\section{Data Preparation}\label{sec:data-prep}

We commence this section by introducing how the dataset is built, going through the photographing conditions; and, the labeling criteria and process.
We specifically note that the dataset is not compiled by ourselves but by the competition organizer.
Subsequently, to ensure the data quality often crucial for machine tasks, we perform a series of preprocessing techniques, ranging from basic ones like resizing the input images and scaling the pixel values, to adopting the Mask R-CNN \cite{he2017mask} to remove noisy backgrounds in the images. Lastly, to remedy the somewhat small amount of data and reduce overfitting, we apply several data augmentation strategies in a random manner during training.

\begin{table}[]
\caption{Grade distribution of AICUP2020. The dataset is roughly balanced, with only slightly more samples in grade B.}\label{tab:dataset_distribution}
\begin{tabularx}{\linewidth}{L|ccc}
\toprule
{Dataset split} & \multicolumn{3}{c}{Grade} \\
& A & B & C \\
\midrule
\textbf{Training set} &  1536 (32.0\%)  & 1786 (37.2\%) & 1478 (30.8\%)\\ 
\textbf{Validation set} & 256 (32.0\%)   & 282 (35.3\%) & 262 (32.8\%) \\
\textbf{Test set}       & 243 (30.4\%)   & 293 (36.6\%) & 264 (33.0\%)\\
\bottomrule
\end{tabularx}
\end{table}

\subsection{The Dataset}\label{subsec:dataset}

The Taiwan AI CUP 2020 competition organizer compiles a dataset of thousands of labeled Irwin mango (a mango cultivar) images for the preliminary round of the contest. For brevity, we refer to the dataset used in the competition as AICUP2020 in our study. 

The AICUP2020 dataset contains 6,400 mango images of varying quality, out of which 4,800, 800, 800 are in training, validation, and test set respectively. The mangoes are held by the collectors and photographed against various backgrounds under different lighting conditions. Each mango is classified into grade A, B, or C based on the evenness of color and severity of defects or diseases. Some samples from the dataset are presented in Figure \ref{fig: sample-imgs}. The grade distribution of AICUP2020 is shown in Table \ref{tab:dataset_distribution}, from which we may see that the data is roughly balanced, with slightly more samples labeled as grade B. The whole dataset is labeled by multiple judges, while each sample is annotated by only one judge.

\subsection{Basic Processings}\label{subsec:basic-process}

\textbf{Image size.}
The input to our models is an RGB image resized to 224x224 pixels. We also attempted to input smaller images (in the hope of saving computation effort) such as 128x128, or even 64x64 ones.
However, the result does not turn out well. 
Due to interpolation, some tiny defects critical for grading would disappear in the resizing process. 
Therefore, we decide to stick with the input size 224x224.

\textbf{Feature scaling.}\label{subsec:value} 
In the raw images, the value of each pixel of a channel lies from 0 to 255, which could hinder the model's convergence if left as is \cite{lecun2012efficient}. Hence, we consider 3 different measures for feature scaling, listed as follows:
\begin{enumerate}
\item \textbf{Simple shifting and scaling}: We transform each pixel $x$ by $x' = (x - 127.5) / 255$. This step keeps the resulting pixel value $x'$ within --0.5 and 0.5, making the training faster and more stable.
\item \textbf{Normalization w.r.t. our dataset}: We compute the RGB mean and standard deviation of the training dataset, and use them to normalize the images such that the resulting data has zero mean and unit variance. 
\item \textbf{Normalization for pretrained models}: As suggested by \texttt{torchvision} guidelines, when fine-tuning the pretrained models, the RGB channels should be normalized with mean=[0.485, 0.456, 0.406] and standard deviation=[0.229, 0.224, 0.225].
\end{enumerate}
For non-pretrained models, the input values are processed by the first method. For pretrained models, the third method is used. We eventually drop the second method since, by our experiment, it leads to the worst performance.

\subsection{Background Removal}\label{subsec:bg-removal}
We attempt 2 methods to achieve the goal; one is the non-learning-based Canny edge detection algorithm \cite{canny1986computational}, and the other is Mask R-CNN \cite{he2017mask}. We find that Canny edge detection segmentation only performs well on a small portion of data with simple backgrounds; thus, we adopt Mask R-CNN as our final solution.
Mask R-CNN is an enhanced version of Faster R-CNN \cite{ren2015faster}, both being robust methods for object detection and image segmentation. Faster R-CNN utilizes Region Proposal Network (RPN) and Region of Interest Pooling (RoIPool) to achieve fast classification and bounding-box regression. Besides the two stages in Faster R-CNN, Mask R-CNN adds a branch parallel to RoIPool for background/foreground discrimination, which predicts whether each pixel is part of an object. Hence, the loss function for the Mask R-CNN consists of 3 components, the classification loss, the bounding-box regression loss, and the binary mask loss. Thanks to the sophisticated network and loss function design, the Mask R-CNN performs well in removing backgrounds for our task. 

We make use of the open-source codes\footnote{\url{github.com/matterport/Mask_RCNN}} for Mask R-CNN in our task. The following are the steps we perform:
\begin{enumerate}
\item We annotate our dataset. 100 images are annotated, of which 60 are used as training data and 40 are kept as validation data.  Although ImageNet does have a ``mango" category, it gives unsatisfactory segmentation results on our data. Hence, we add the ``Irwin mango" category and utilize the VGG Image Annotator \cite{dutta2019via} to mark the mangoes' positions with polygons for further fine-tuning.
\item We fine-tune Mask R-CNN on the 60-image training set. We initialize the model with ImageNet pretrained weights. We assume the first few layers of the network are already well-trained to extract low-level features, hence we freeze their weights and only allow the last layers to be updated. The best result is achieved with 20 fine-tuning epochs and learning rate 1e--3.
\item We perform image segmentation. At first, we use a splash method to extract mangoes from images, i.e., finding the exact boundary of the mango. However, our classification models perform not as desired with these data, most likely due to the rugged outline of the extracted mangoes. Thus, we finally use the bounding box method. The bounding box is obtained from the extreme points of the border given by the splash method. We find that entire mangoes can be better preserved with bounding boxes. 
\end{enumerate}

\begin{figure}[h]
 \centerline{
 \includegraphics[trim={0 0 0 0},clip,width=\linewidth]{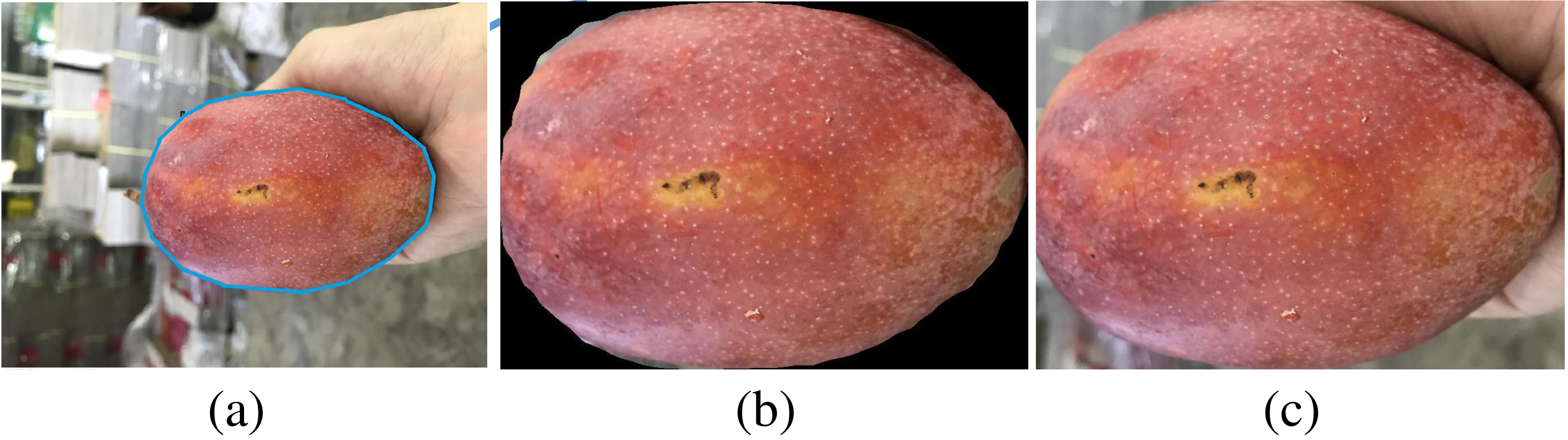}}
 \caption{Background removal process. Image (a) shows how we annotate our dataset; the boundary of the mango fruit is marked with a blue polygon. As for segmentation results, Image (b) uses splash method, with which a rugged boundary of the mango is obtained. Image (c) uses the bounding box method, with which the whole mango is preserved.}
 \label{fig:mask_rcnn_samples}
\end{figure}
% \vspace{-5mm}
\subsection{Data Augmentation}\label{subsec:data-aug}
To obtain the highest possible amount of data augmentation, in each training epoch, we randomly apply the following set of perturbations to every training sample:
\begin{itemize}
\item \textbf{Horizontal or vertical flip}, each with 50\% probability;
\item \textbf{Brightness}, --20 to +20\%;
\item \textbf{Contrast}, --10 to +10\%;
\item \textbf{Rotation}, --20 to 20 degrees;
\item \textbf{Zoom in/out}, 0.8x to 1.25x.
\end{itemize}
By our experiment, data augmentation drastically lessens overfitting with little negative effect on training time.

\section{Single-Task CNNs}\label{sec:e2e-cnn}

Owing to their great achievements on the ImageNet corpus, we adopt the following well-known CNNs for grading the mangoes: AlexNet, VGG11 (w/ batchnorm), VGG16 (w/ batchnorm), and ResNet34. This section introduces the adopted CNNs and lays out the details for training (or fine-tuning) them.

% We try the following CNN models:
% \begin{itemize}[noitemsep,topsep=0pt]
% \item AlexNet
% \item A model with VGG backbone with small variations (our own implementation)
% \item VGG11 with batch normalization
% \item VGG16 with batch normalization
% \item ResNet34

% \end{itemize}
% In order to find out which end-to-end model brings the best accuracy, we first implement our own model and compare it with a VGG16 without pretrained weights. After that, we compare the train-from-scratch VGG16 model with above-mentioned models pre-trained on ImageNet. 

\subsection{Network Introduction}\label{subsec:end2end-archi}
\textbf{AlexNet} \cite{krizhevsky2012imagenet} is the very first successful CNN on the ImageNet dataset. It contains 5 convolutional layers and 3 fully-connected layers. Dropout, ReLU nonlinearity, and max-pooling are also present in the architecture.

\textbf{VGG}s \cite{simonyan2014very} improve over AlexNet by replacing large-sized convolutional kernels with multiple 3x3 and 1x1 kernels; and, increasing the network depth. A VGG has 5 “VGG blocks”, which are composed of a sequence of convolutional layers, each followed by ReLU nonlinearity, and a max-pooling layer. VGG11 and VGG16 are named according to the number of convolutional and fully-connected layers used in the model.

\textbf{ResNet} \cite{he2015deep} utilizes skip connections to solve the \textit{gradient vanishing} problem often encountered when training deeper neural networks. ResNet34 has 4 sub-modules which consist of 3, 4, 6, and 3 basic blocks respectively. A basic block is composed of 2 convolutional layers with batch normalization and ReLU activation; and, a connection that bypasses the 2 layers. It is called ResNet34 for having 33 convolutional layers and 1 fully-connected layer.

\subsection{Training Details}\label{subsec:end2end-train}
All networks mentioned above are trained via minimizing the multi-class cross-entropy loss:
\begin{equation}\label{eqn:multi-ce}
   \mathcal{L} = - \frac{1}{N}\displaystyle \sum_{i=1}^{N} \log{p(y_i)}; \; y_i \in \{\text{A, B, C}\},
\end{equation}
where $N$ is the number of training samples, and $p(y_i)$ is the (softmax-ed) probability the network assigns to the sample's ground-truth class. Some of the hyperparameters and training settings, listed in the following paragraphs, are hand-picked by us to achieve the best performance.

For non-pretrained models, the weights in every layer are initialized with He uniform initialization \cite{he2015delving} and the bias is filled with 0. We choose batch size 32, and Adam optimizer with learning rate set to 1e--4.

For pretrained models, we initialize the model with ImageNet pretrained weights, and replace the last fully-connected with a new one for our task (output dimension=3). The last layer's weights are initialized with He uniform initialization and the bias is filled with 0. We train the models using stochastic gradient descent (SGD) with batch size 32 and momentum 0.9. The learning rate is initially set to 1e--3, and then decreased by 90\% every 15 epochs. To reduce overfitting, we apply a 50\% dropout rate to the fully-connected layers. 

We use early-stopping as regularization and termination condition. The training is terminated if the validation accuracy has not improved for 20 epochs. 

\section{Convolutional Autoencoder-Classifiers}\label{sec:conv-ae}

% In addition to end-to-end CNN classifiers, we approach this task with another network structure, which involves:
% \begin{itemize}[topsep=0pt, noitemsep]
%     \item A convolution-based \textbf{encoder} that compresses an image into a latent vector;
%     \item A convolution-based \textbf{decoder} that reconstructs the image from the latent vector and some intermediate encoded features;
%     \item A fully-connected \textbf{classifier} that takes the latent vector as input and gives the class prediction.
% \end{itemize}
% The rationale behind this solution is as follows: first, the presence of autoencoder forces the network to preserve the essential information for reconstruction, rather than solely cramming for the prediction task, thereby reducing the overfitting of the classifier; second, the compressed latent features also provide more flexibility for downstream tasks, for example, if we prefer doing a ``type of defect'' classification instead, we can keep the architecture and weights of the autoencoder intact, replace the classifier part with a task-specific one, and simply fine-tune the network for the task. 
% In this sense, whatever the prediction task is, the network can take advantage of what it has learned from all the mango pictures it's seen, instead of being confined to a single task.

In addition to the aforementioned single-task CNNs, being inspired by the auspicious attempts to adopt multi-task learning for classification tasks \cite{zhang2016augmenting, le2018supervised}, we present another series of models that contain an additional autoencoder part for reconstructing the input image, hence their name ``convolutional autoencoder-classifiers'' (or, ``ConvAE-Clfs'' for short).
This section, similar to the previous one, introduces the architecture of the networks we use and lists the training details.

\subsection{Network Introduction}\label{subsec:conv-ae-archi}
The ConvAE-Clfs consist of 3 components:
\begin{itemize}
    \item A convolution-based \textbf{encoder} that compresses an image into a latent vector;
    \item A convolution-based \textbf{decoder} that reconstructs the image from the latent vector and some intermediate features;
    \item A fully-connected \textbf{classifier} that takes the latent vector as input and gives the class prediction.
\end{itemize}
Our implementation of ConvAE-Clfs is based on the open-source codes\footnote{\url{github.com/ternaus/angiodysplasia-segmentation}} for the networks presented in a previous work on angiodysplasia (an intestinal disease) detection \cite{shvets2018angiodysplasia}. In that work, 3 encoder-decoder architectures were proposed, with the main difference lying in their pretrained encoders:
\begin{itemize}
    \item \textbf{TernausNet11}---contains VGG11 encoder;
    \item \textbf{TernausNet16}---contains VGG16 encoder;
    \item \textbf{AlbuNet34}---contains ResNet34 encoder.
\end{itemize}

\begin{figure*}[h]
    \centering
    \includegraphics[width=\textwidth, trim={0, 3mm, 0, 3mm}]{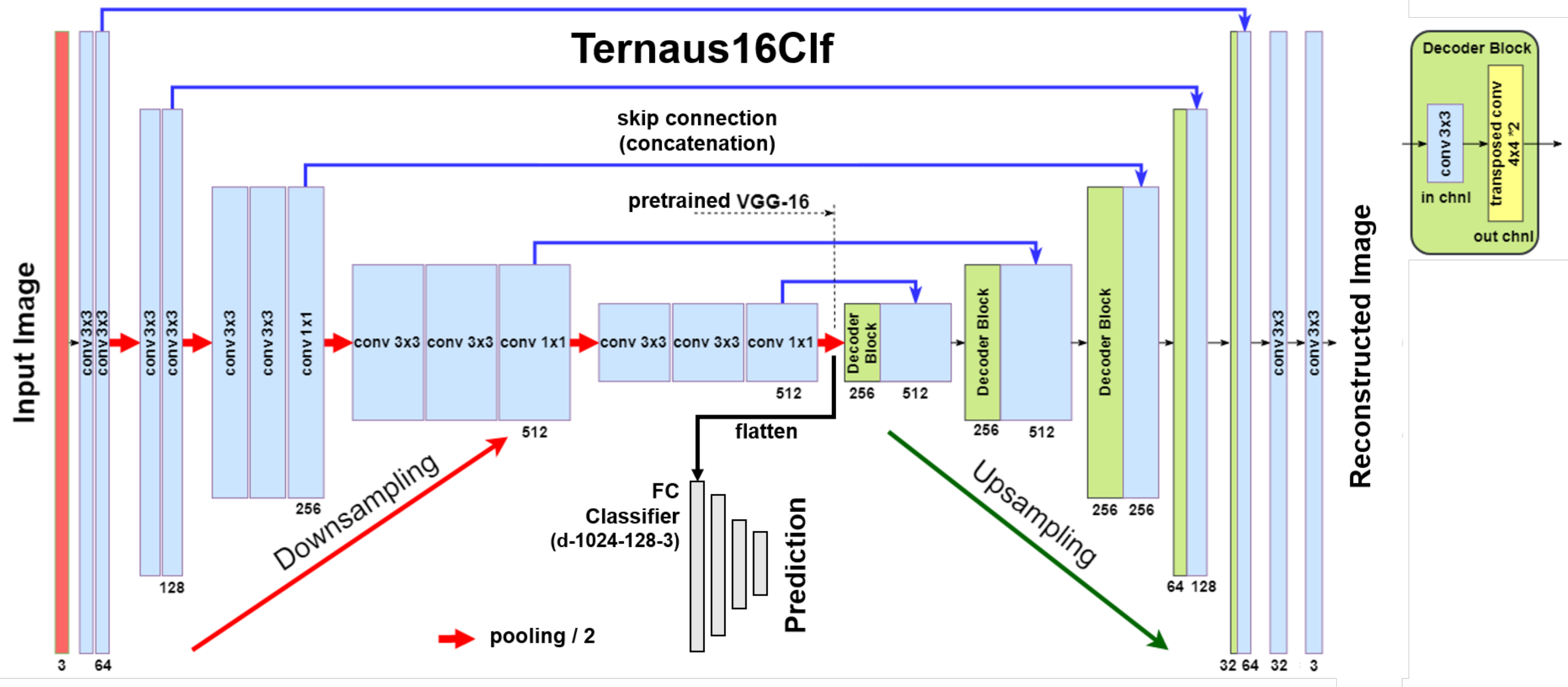}
    \caption{The architecture of \textbf{Ternaus16Clf}. This figure is modified from the angiodysplasia detection paper \cite{shvets2018angiodysplasia}.}
    \label{fig:ternaus16}
% \vspace{-0.3cm}
\end{figure*}

We revamp the networks to suit our classification task and dub them \textbf{Ternaus11Clf}, \textbf{Ternaus16Clf}, and \textbf{Albu34Clf} respectively. Figure \ref{fig:ternaus16} is a schematic of the \textbf{Ternaus16Clf}'s architecture (the other 2 networks are similarly structured). For each convolutional block in the encoder, there is a corresponding decoder deconvolutional block in charge of reconstruction, which takes its input not only from its preceding block, but also from a skip connection linked to an encoder convolutional block.
Working in alongside the decoder is the fully-connected, LeakyReLU-activated classifier of dimensions $d$-$1024$-$128$-$3$ for each layer, where $d$ is the dimension of latent features received from the encoder.

\subsection{Training Details}\label{subsec:conv-ae-training}

Since the networks contain both an autoencoder and a classifier, a hybrid loss is required for optimization; one part of which is the reconstruction loss:
\begin{equation}\label{eqn:mse}
    \mathcal{L}_{rec} = \frac{1}{N} \displaystyle \sum_{i=1}^N || \hat{\mathbf{x}}_i - \mathbf{x}_i ||_2^2,
\end{equation}
which is the mean squared error (MSE) between the reconstructed image $\hat{\mathbf{x}}_i$ and the input image $\mathbf{x}_i$, and the other part is the classification loss $\mathcal{L}_{clf}$, which is the same loss (see Eqn. \ref{eqn:multi-ce}) applied to single-task CNNs. 
The hybrid loss is obtained via $\mathcal{L} = \alpha \mathcal{L}_{rec} + (1-\alpha) \mathcal{L}_{clf}$, where $\alpha$ is set beforehand.
We keep the skip connections between encoder and decoder since we find them having a stabilizing effect on training, i.e., with them, the classification accuracy fluctuates less severely.
Through experiments, we find the autoencoder part quite robust, probably due to the encoder's pretrained weights and the decoder's access to intermediate features via skip connections; therefore, we set $\alpha$=0.05 for a balanced optimization.

We choose batch size 64, Adam optimizer with initial learning rate 1e--4, and the learning rate scheduler ``ReduceLROnPlateau'', which decays the learning rate by 80\% whenever the validation accuracy hasn't improved for 8 epochs. 
The training process is terminated by early-stopping with 20 epochs of patience on the improvement of validation accuracy.
Furthermore, a 40\% dropout rate is applied to all layers of the fully-connected classifier.
The entire training process, for the 3 networks alike, takes about 1.5 hours on an NVIDIA V100 GPU with 32GB memory.

\section{Experiments and Discussions}\label{sec:exp}

In this section, we present and compare the performance of our proposed models, and discuss the effectiveness of using pretrained weights, adopting Mask R-CNN for background removal, and augmenting an autoencoder to the model. 
Moreover, through saliency maps \cite{Simonyan2014deep} and PCA \cite{hotelling1933analysis}, we provide some insights into some of both correctly-classified and misclassified samples by VGG16, the famed and best-performing architecture in our work.
We note that all of the model checkpoints put to test are those scoring the highest validation accuracy during training, and that data augmentation (see Sec. \ref{subsec:data-aug}) is applied to every model involved in this section.

% In the previous section, we present the details of model configurations and training parameters. In this section, we would describe the details of model evaluation.
% \shihlun{Will write a section introduction later.}

\subsection{Single-Task CNNs}\label{subsec:end2end-exp}
\begin{table}[h]
\caption{Accuracies of single-task CNNs. \textbf{bg. rem.} means Mask R-CNN (refer to Sec. \ref{subsec:bg-removal}) is adopted to remove the irrelevant backgrounds of the training images.}\label{tab:end2end-acc}
\begin{tabularx}{\linewidth}{L|ccc}
\toprule
Model & \multicolumn{3}{c}{Accuracy} \\
& Train & Val. & Test \\
\midrule

\textbf{non-pretrained VGG16}     & 80.7 \%   &  81.4 \% & 80.6 \%\\ \hline
\textbf{pretrained AlexNet}       & 86.5 \%   & 80.4 \% & 80.4 \%\\
\textbf{pretrained VGG11}         & 85.8 \%   & 81.8  \% & 79.3 \%\\
\textbf{pretrained VGG16}         & 88.1 \%   & 82.6 \% & \textbf{82.8} \%\\
\textbf{pretrained ResNet34}      & 88.1 \%   & 79.6 \% & 81.1 \%\\ \hline
\textbf{pretrained AlexNet w/ bg. rem.}     & 81.9 \%   & 81.0 \% & 81.3 \%\\
\textbf{pretrained VGG11 w/ bg. rem.}      & 84.8 \%   & 81.8 \% & 81.9 \%\\
\textbf{pretrained VGG16 w/ bg. rem.}      & 87.4 \%   & 83.1 \% & \textbf{83.5} \%\\
\textbf{pretrained ResNet34 w/ bg. rem.}      & 87.5 \%   & 80.5 \% & 82.4 \%\\
\bottomrule
\end{tabularx}
\end{table}

\textbf{With or without pretrained weights.} From Table \ref{tab:end2end-acc}, we can see that the VGG16 initialized with pretrained weights performs better than the non-pretrained one, with a gain of 2.2\% on test accuracy.
Also, it takes significantly less time to fine-tune pretrained models; it takes 2 hours to train from scratch, while fine-tuning only takes about half an hour.
Therefore, we suppose transfer learning is a more efficient and efficacious way than training from scratch, and only consider pretrained models in subsequent experiments.

\textbf{Different pretrained models.} After trying different models, we find VGG16 performing the best (see Table \ref{tab:end2end-acc}) and also an easier one to tune. In the case where Mask R-CNN background removal is not applied, the pretrained VGG16 outperforms the runner-up, ResNet34, by 1.7\%.

\textbf{With or without Mask R-CNN background removal.} Results in Table \ref{tab:end2end-acc} show that training with images having irrelevant background removed leads to higher validation and test accuracy than with the original images. The improvement is noticeable for all models involved, with the gain in test accuracy ranging from 0.7\% to 2.6\%. This is probably due to that the model need not learn to focus on the mangoes by itself and that the resolution of the mangoes is higher after resizing. Besides, the VGG16 is still the best performer with Mask R-CNN background removal.

\subsection{Convolutional Autoencoder-Classifiers}\label{subsec:exp-conv-ae}
\begin{table}[]
\caption{Performances of the convoltional autoencoder-classifiers, and the direct comparison with their single-task CNN counterparts. Reconstruction loss is measured by mean squared error (MSE). All models are initialized with pretrained weights and trained with the data after Mask~R-CNN background removal.}\label{tab:conv-ae-res}
\begin{tabularx}{\linewidth}{L|ccc|ccc}
\toprule
Model & \multicolumn{3}{c|}{Reconstr. loss} & \multicolumn{3}{c}{Accuracy} \\
& Train & Val. & Test & Train & Val. & Test\\
\midrule
\textbf{VGG11} & \textit{n.a.}   & \textit{n.a.} & \textit{n.a.} & 84.8 \%   & 81.8  \% & 81.9 \% \\
\textbf{Ternaus11Clf}     & .003   & .003 & .003 & 92.6 \% & 83.9 \% & \textbf{82.6} \% \\ \hline
\textbf{VGG16} & \textit{n.a.}   & \textit{n.a.} & \textit{n.a.} & 87.4 \%   & 83.1  \% & 83.5 \%  \\
\textbf{Ternaus16Clf}     & .006 & .007 & .007 & 90.5 \% & 84.8 \% & \textbf{83.6} \% \\ \hline
\textbf{ResNet34} & \textit{n.a.} & \textit{n.a.} & \textit{n.a.} & 87.5 \% & 80.5 \% & \textbf{82.4} \% \\
\textbf{Albu34Clf}     & .040  & .029 & .028 & 86.2 \% & 81.9 \% & 81.4 \% \\
\bottomrule
\end{tabularx}
\end{table}
% \vspace{-3mm}
Table \ref{tab:conv-ae-res} displays the reconstruction and classification performances of the convolutional autoencoder-classfiers (see Sec. \ref{sec:conv-ae}) trained on background-removed images and initialized with pretrained weights.
The numbers indicate that the networks containing VGG encoders, i.e., Ternaus11Clf and Ternaus16Clf, compared to the ResNet34-based Albu34Clf, are not only better autoencoders but also stronger classifiers.

Comparing these networks with single-task CNNs (also see Table \ref{tab:conv-ae-res}), the Ternaus11Clf and Ternaus16Clf achieve higher validation accuracy than their single-task VGG counterparts, while performing comparably or slightly better on the test set. 
On the other hand, the Albu34Clf falls behind ResNet34 on test accuracy; we suspect that this is due to the relatively poor reconstruction ability in the first place, i.e., less vital information is encoded in the latent features.
Plus, it is noticeable that the ConvAE-Clfs suffer a performance gap between the test set and validation set (0.5\% to 1.3\%); however, the cause is still yet to be discovered.

All in all, the ConvAE-Clfs do not possess an advantage in our task. Nevertheless, this result should be taken with a pinch of salt as our test data is small in size (800 images); and, we are yet to deploy the trained autoencoders to related tasks, such as a ``type of defect'' classification, to examine whether they could be beneficial.

% we can also observe an incremental performance gain in classification; however, this improvement should be taken with a pinch of salt since our validation data is small (800 pictures) and only few rounds of experiments are run. 
% Besides, we also notice that during training, the accuracy fluctuates more severely compared to end-to-end CNNs. We suspect that this is due to some competition between the reconstruction and classification losses.
% the difference in training accuracy should not be over-explained, for it largely depends on training settings (e.g., \# epochs, learning rate schedules, etc.).

\subsection{Insights into the Models’ Decisions}\label{subsec:insights-model}

\begin{figure}[]
 \centerline{
 \includegraphics[trim={0 0 0 0cm},clip, width=\linewidth]{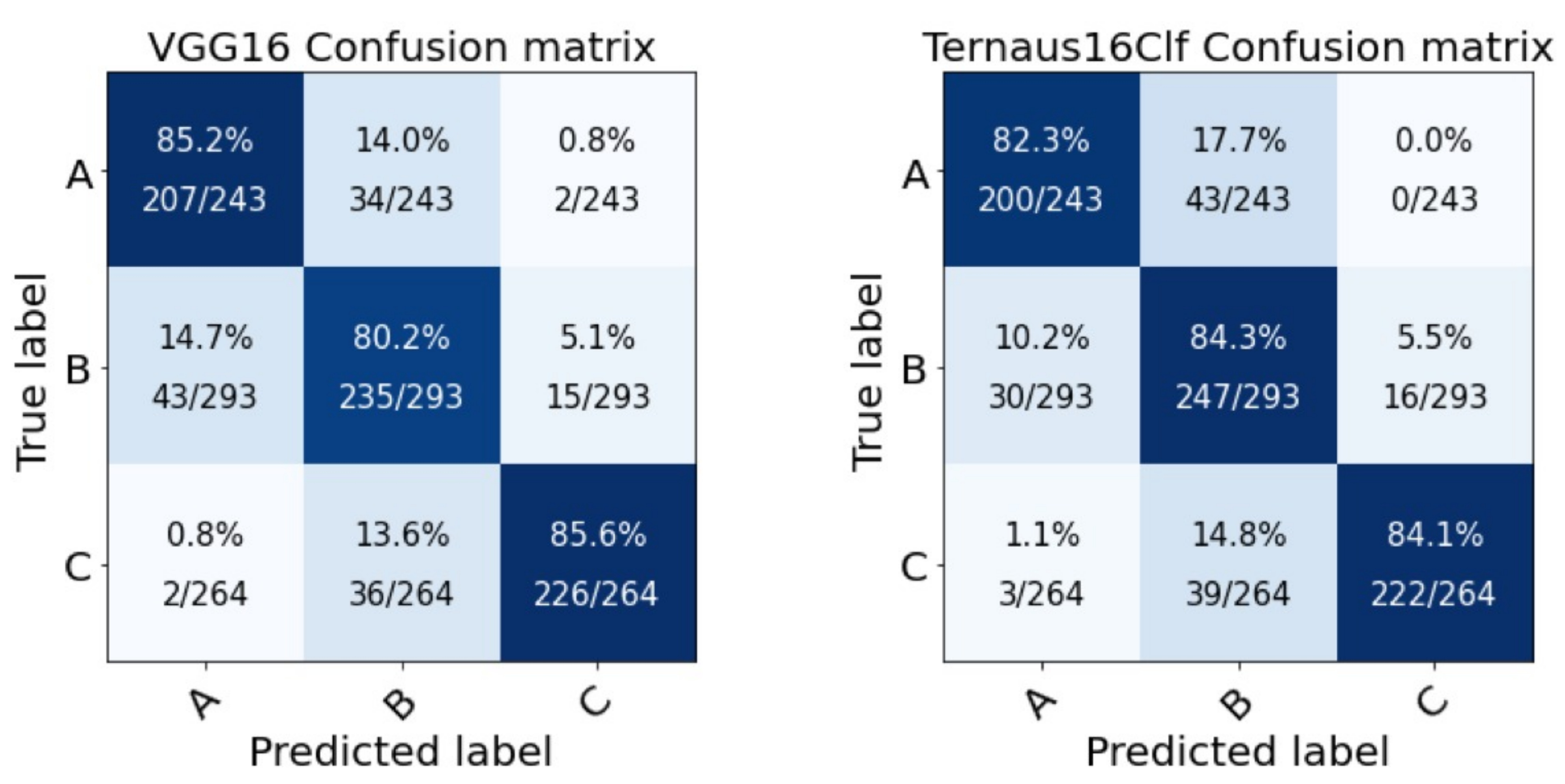}}
\caption{The confusion matrices of the pretrained VGG16 and Ternaus16Clf computed on the test set. The denominators indicate the number of samples in the grade (ground truth).}
 \label{fig: cm}
\end{figure}

\textbf{Confusion matrices.} Since VGG16 and Ternaus16Clf outperform other models in our experiments, we decide to take a closer look into their predictions. From the confusion matrices (Figure \ref{fig: cm}), we can observe that it is harder for both models to tell apart grade A and grade B mangoes. Also, grade C mangoes are often misclassified as grade B.

As VGG16 is a tested-and-proved model in many deep learning image classification problems, we especially desire to dig into how it works in our task. We utilize saliency maps and PCA to take a deeper look into the VGG16. Please note that the two techniques can also be easily applied to other models.

\textbf{Saliency maps.} 
The saliency map \cite{Simonyan2014deep} visualizes the model's attention on the input image when making prediction on it. A saliency map $\mathcal{S}$ is of the same size as the input image, with each of its pixels $\mathcal{S}_{ij}$ obtained by:
\begin{equation}\label{eqn:saliency-map}
    \mathcal{S}_{ij} = \norm { \frac{\partial p(\hat{y}_{\mathbf{x}})}{\partial \mathbf{x}_{ij}} }_2,
\end{equation}
which is the size of the gradient of the (softmax-ed) probability of the predicted class $p(\hat{y}_{\mathbf{x}})$, with respect to the input pixel $\mathbf{x}_{ij}$. Note that we use vector norm since each pixel consists of RGB channels. The intuition is that the pixels contributing larger gradient are more significant to the model's decision.

\begin{figure}[]
 \centerline{
 \includegraphics[trim={0 0 0 0},clip, width=\linewidth]{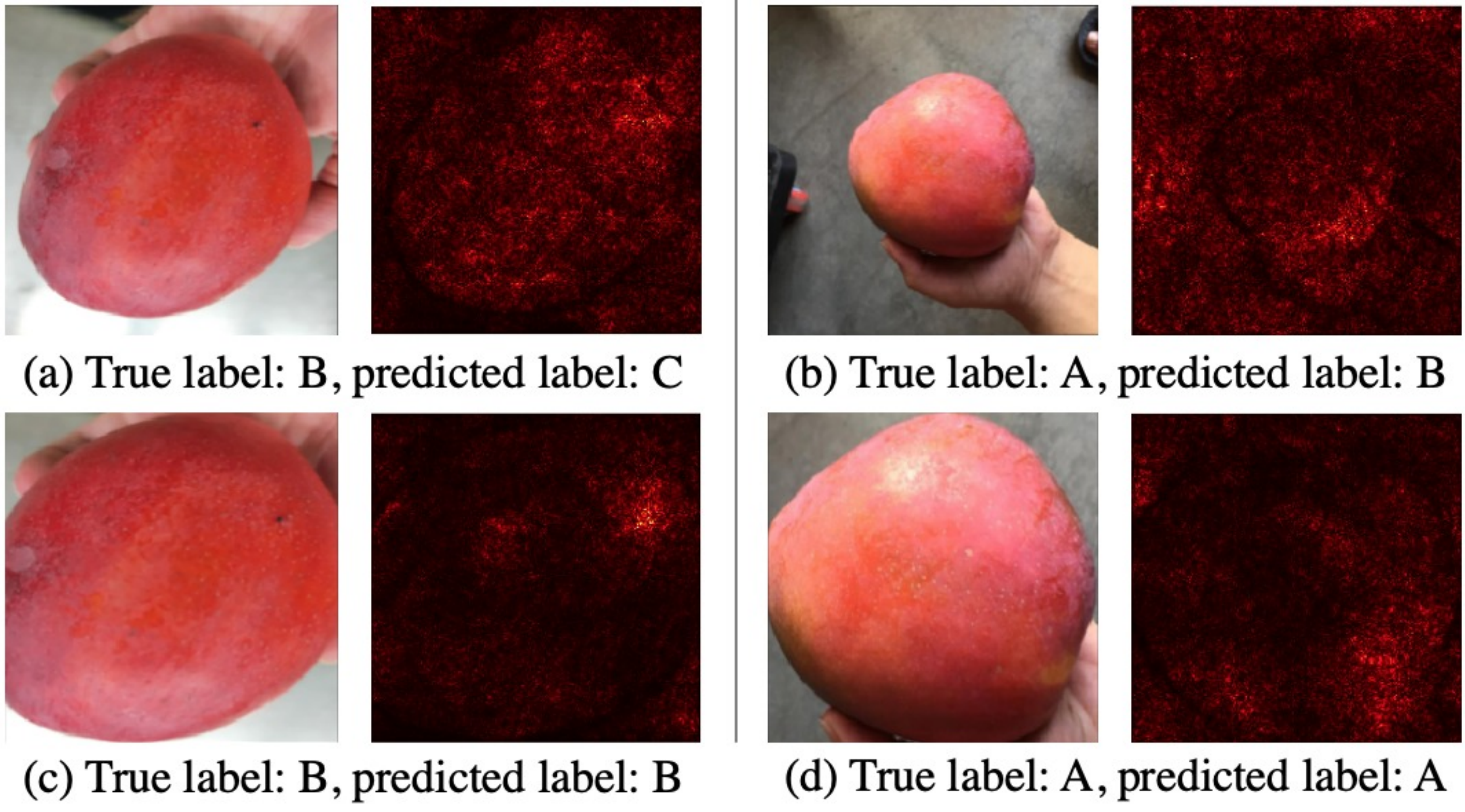}}
% \caption{(a) and (b) are results from VGG16 trained with original images; (c) and (d) results from VGG16 trained with background removal. The saliency maps \cite{Simonyan2014deep} next to each samples visualize the attention of VGG16. Red pixels in the saliency maps indicate large gradients w.r.t. the predicted classes.}
\caption{Saliency maps of the VGG16 on 2 test set samples, comparing the cases where background removal is not applied ((a), (b)) or applied ((c), (d)). Please note that (a) and (c) are the same samples in the dataset, and so are (b) and (d). Red pixels in the saliency maps indicate large gradients w.r.t. the predicted classes.}
 \label{fig: compare_bg_rem}
\end{figure}

First, we would like to know why training with background-removed images leads to better performance. In Figure \ref{fig: compare_bg_rem}, we present 2 samples on which VGG16 trained with original images makes mistakes, while VGG16 trained with background-removed images classifies correctly. 
From Figure \ref{fig: compare_bg_rem} (a) and (b), we speculate that the model makes wrong decisions for being distracted by the irrelevant background; from Figure \ref{fig: compare_bg_rem} (c) and (d), we can see that the model pays more attention to the mangoes and makes the right prediction after applying background removal.

\begin{figure}[]
 \centerline{
 \includegraphics[trim={0 0 0 0},clip, width=\linewidth]{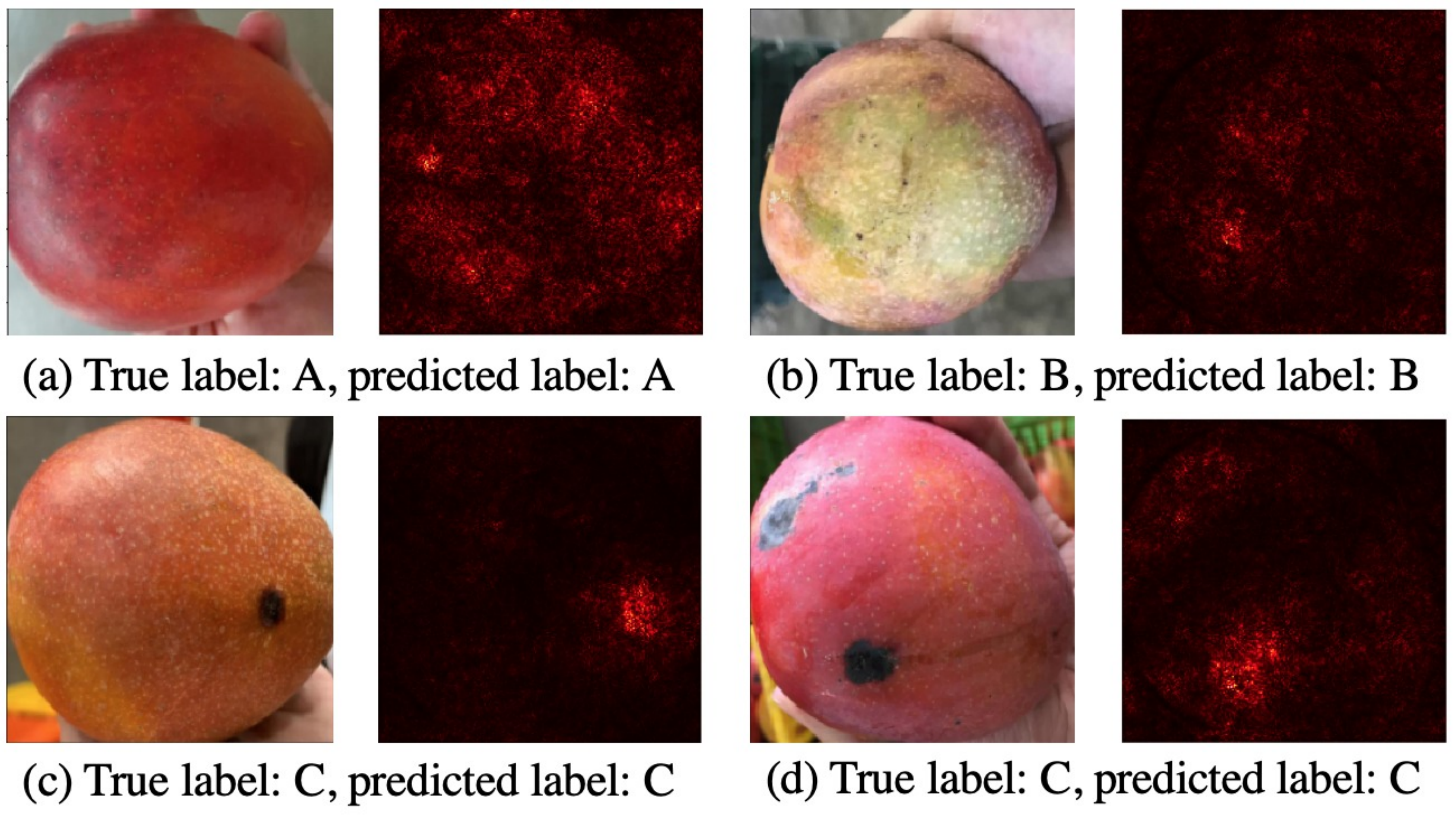}}
%  \vspace{-0.2cm} 
\caption{Correctly-classified samples by the VGG16 and their corresponding saliency maps. The samples show that the model puts appropriate attention on the mangoes' defects.}
 \label{fig: correctly-classified}
\end{figure}

\begin{figure}[]
 \centerline{
 \includegraphics[trim={0 0 0 0},clip, width=\linewidth]{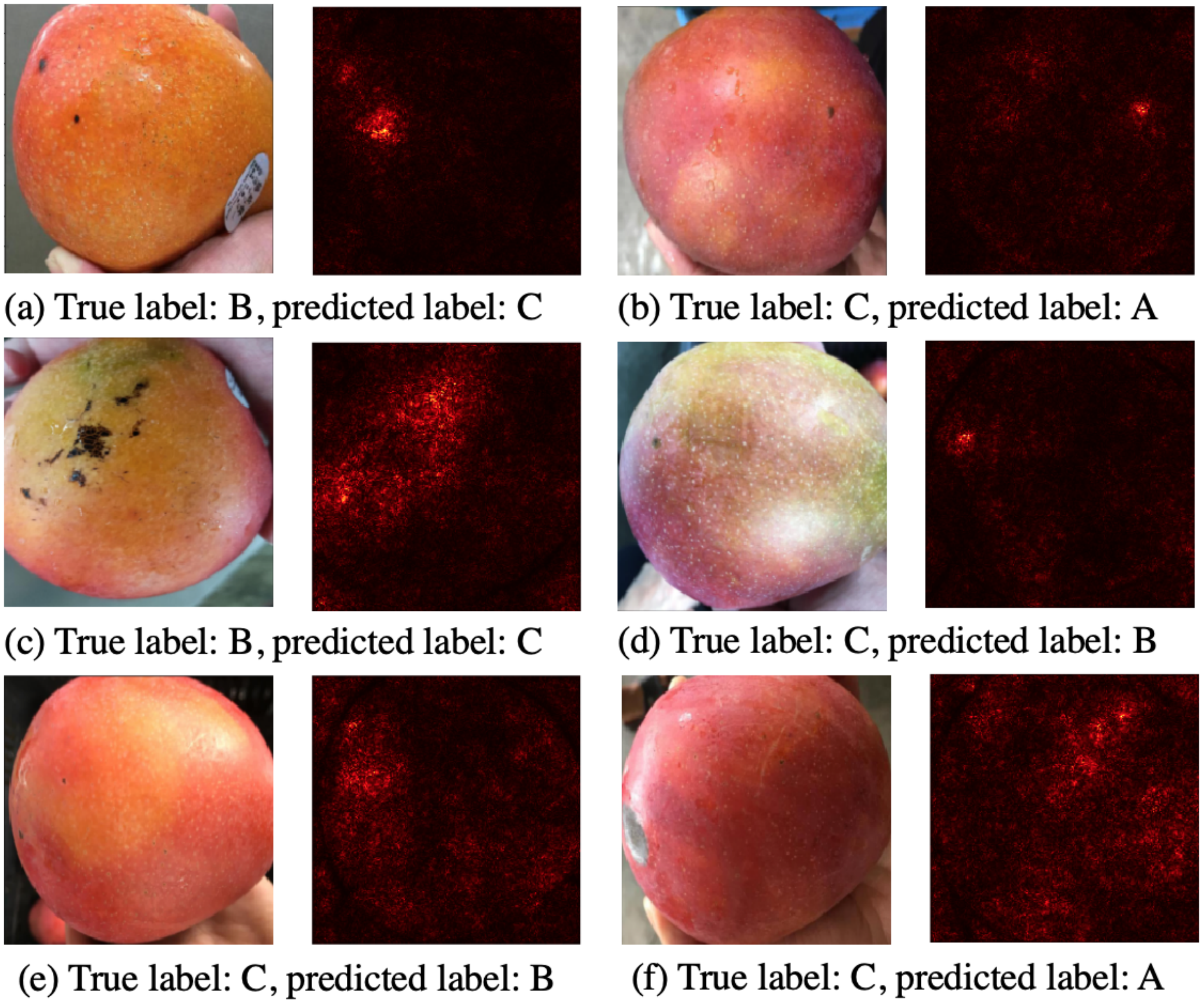}}
%  \vspace{-0.2cm} 
\caption{Misclassified samples by the VGG16 and their corresponding saliency maps. The model's attention could be justified as it still focuses on the black dots when they are present, or the entire mango when there are no clearly visible defects.}
 \label{fig: wrongly-classified}
\end{figure}

Next, we strive to find out the weaknesses of VGG16 trained on the dataset with removed backgrounds. Figure \ref{fig: correctly-classified} presents some examples that are correctly classified by the model, from which we can see that the model does focus on the mangoes, more importantly, on the defects that affect their quality grade. Then, we sort the misclassified mangoes by their cross-entropy loss values. Figure \ref{fig: wrongly-classified} displays some of the samples with higher loss. We can observe that, consistent with the correctly-classified samples, the model puts most of its attention on the mangoes and defects like black dots.

Looking at the misclassified mangoes, we find the model makes mistakes on samples involving uneven skin colors more often, as can be seen from Figure \ref{fig: wrongly-classified}(d); the model is not quite aware of the color variation on the mango's upper-right corner. 
Other than that, the model's attention seems to be reasonable. 
In the process of examining these samples, we find the labeling standard quite inconsistent. 
For instance, some mangoes with uneven colors are labeled as A, while some are labeled as B; plus, some samples with only tiny defects are labeled as C (e.g., Figure \ref{fig: wrongly-classified}(c)).
This can result from the fact that each sample is annotated by only one person; hence, we suggest that the labels be cross-checked by the annotators.

\textbf{PCA on the latent feature vectors.}
In order to realize how the model perceives the input images in its latent feature space, we leverage principal components analysis (PCA) \cite{hotelling1933analysis} to explain the extracted features after the convolutional layers. PCA is a method to reduce high-dimensional feature vectors into low-dimensional ones through eigendecomposition on the dataset, and projecting the original vectors onto the maximum-variance eigenvectors, i.e., the principal components; the coefficients (associated with projection) of these components represent the most substantial differentiating attributes of the samples in the model's eyes. 

\begin{figure}[]
\centering{  
\includegraphics[trim={0 0 0 0},clip, width=\linewidth]{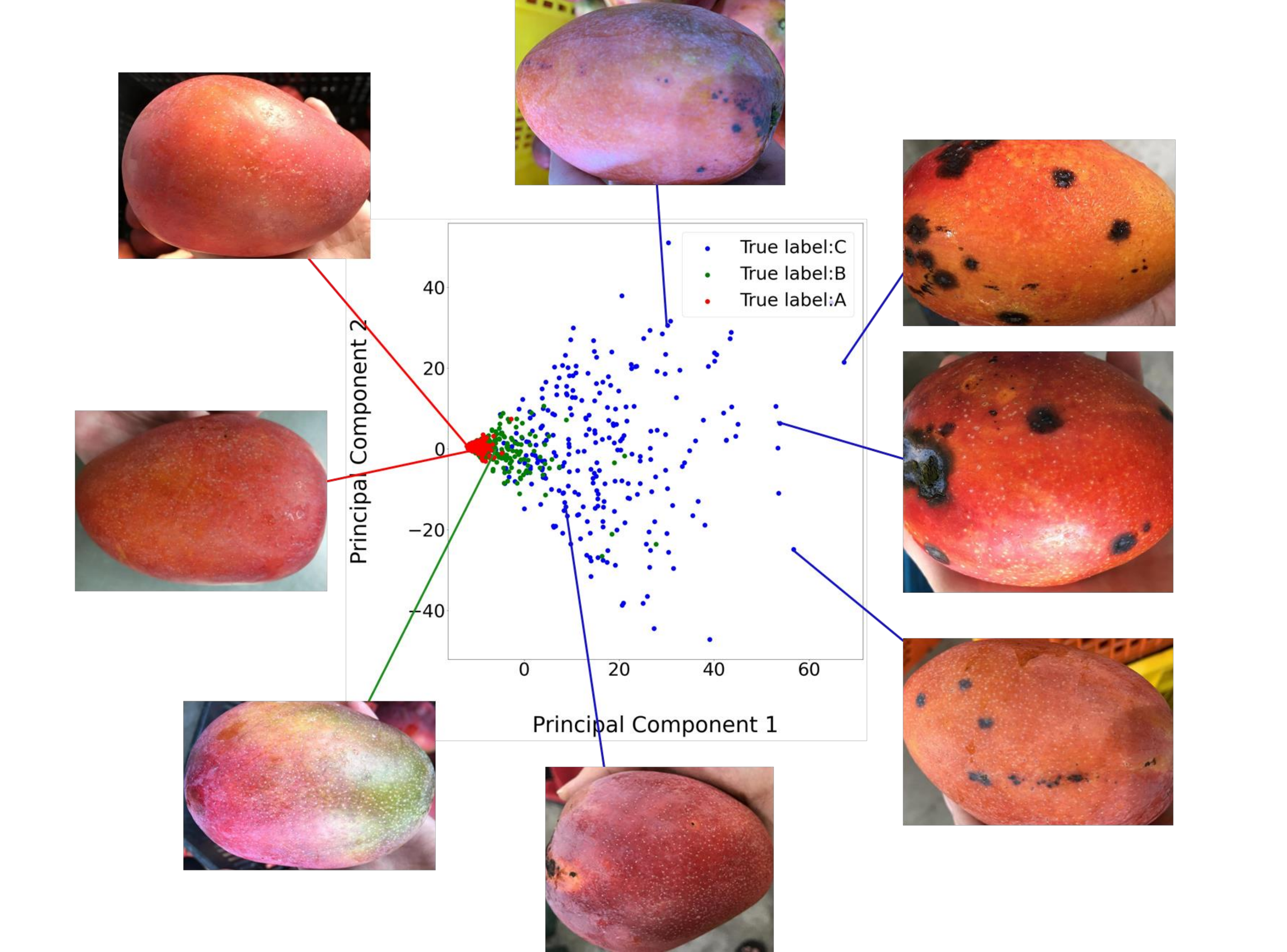}}
\caption{2-component PCA on the latent features of VGG16 and some sample mangoes corresponding to different PCA coefficients. It is evident that the 1st principal component encodes the severity of defects, and largely determines the quality grade, while the focus of the 2nd principal component is rather obscure.}
\label{PCA_result}
\end{figure}

Looking at Figure \ref{PCA_result}, it is obvious that the 1st principal component differentiates most of our data: samples of grade C get the highest coefficients, followed by grade B, then grade A.
The number and area of black spots seem to be the determinant of the value.
Meanwhile, the coefficient of the 2nd principal component only varies greatly on grade C samples, but we couldn't extract its meaning from the values.

Nevertheless, it is worth noting that, on the plot (center, Figure \ref{PCA_result}), grade A samples are the most densely packed, grade B ones less so, and grade C mangoes scatter all over the plane.
This phenomenon potentially explains the nature that all high-quality mangoes are alike, and that the causes of low-quality mangoes are numerous and diversified.

% \vspace{-0.25cm}
\section{Conclusion}\label{sec:conclusion}
In this paper, we have combined and investigated several deep learning-based methods to approach the mango grading problem. Through our experiments, we found that the VGG16 is the best model for the task; and, removing the irrelevant background of images with Mask R-CNN and making use of ImageNet pretrained weights are effective ways to boost the accuracy. The proposed convolutional autoencoder-classifiers were shown to have no clear advantage over the single-task CNNs, but the result should be verified with larger datasets and more related tasks. Furthermore, we provided additional insights into the VGG16's working with saliency maps and PCA. The analyses have demonstrated that the model actually learned to base its decision on the mangoes' visible defects. The explainable insights can also be presented to humans as another layer of assurance when such systems are deployed to real-world scenarios.

\section{Acknowledgement}
We would like to express our sincere gratitude to the \textbf{Behavioral Informatics \& Interaction Computation (BIIC) Lab} (National Tsing Hua University, Hsinchu, Taiwan) for compiling and releasing the AICUP2020 Irwin mango dataset.
\bibliography{reference}
\bibliographystyle{IEEEtran}

\end{document}